%% file: main_review.tex
\documentclass[10pt,twocolumn,letterpaper]{article}

\usepackage{wacv}

\makeatletter
\@namedef{ver@everyshi.sty}{}
\makeatother
\input{preambles/packages.tex}
\input{preambles/definitions.tex}

\wacvfinalcopy 

\def\wacvPaperID{560} 

\ifwacvfinal\fi
\begin{document}


\title{L*ReLU: Piece-wise Linear Activation Functions \\
for Deep Fine-grained Visual Categorization}

\author
{
    Mina Basirat \hspace{2cm} Peter M. Roth \\
    Institute of Computer Graphics and Vision\\
    Graz University of Technology\\
    {\tt\small \{mina.basirat,pmroth\}@icg.tugraz.at}
}

\maketitle
\ifwacvfinal\thispagestyle{empty}\fi

\input{sections/abstract}

\input{sections/introduction}
\input{sections/related}

\input{sections/method}
\input{sections/experiments}

\input{sections/conclusion}

\paragraph{Acknowledgement} This work was partially supported by FFG Bridge project SISDAL (21604365). We gratefully acknowledge the support of NVIDIA Corporation with the donation of the Titan Xp GPU used for this research.

\clearpage

{\small
\newcommand{\BIBdecl}{\setlength{\itemsep}{0 em}}
\bibliographystyle{IEEEtranN}
\bibliography{refs/abbrv,refs/mina,refs/pmroth}
}


\end{document}

%% file: preambles/packages.tex
\usepackage{support-caption} 
\usepackage{graphicx}
\usepackage{nicefrac}
\usepackage{subcaption}   
\usepackage[numbers,sort,compress]{natbib}

\usepackage{tikz}
\usetikzlibrary{decorations.pathreplacing,shapes,positioning,arrows}
\usepackage{pgfplots}

\usepackage{caption,booktabs,array} 

\usepackage{footmisc}   
\usepackage{url}

\usepackage{eso-pic}
\usepackage{times}
\usepackage{epsfig}
\usepackage{amsmath}
\usepackage{amssymb}
\usepackage{flushend}



%% file: preambles/definitions.tex
\newcommand{\func}[1]{\emph{#1}}

\newcommand{\elu}{\func{ELU}\xspace}
\newcommand{\selu}{\func{SELU}\xspace}
\newcommand{\relu}{\func{ReLU}\xspace}
\newcommand{\lrelu}{\func{LReLU}\xspace}
\newcommand{\prelu}{\func{PReLU}\xspace}
\newcommand{\rrelu}{\func{RReLU}\xspace}
\newcommand{\swish}{\func{Swish}\xspace}

\newcommand{\TahX}{\func{Tanh ($\alpha x$)+ $\beta x$}\xspace}

\newcommand{\pswish}{\func{PSwish}\xspace}
\definecolor{BrightRoyalPurple}{rgb}{0.65, 0.05, 0.78}
\definecolor{DarkGreen}{rgb}{0.0, 0.4, 0.0}

\newcommand{\mbrmk}[1]{\textcolor{DarkGreen}{\textbf{MB:} [* #1 *]}}

\newcommand{\comment}[1]{}

%% file: sections/abstract.tex
\begin{abstract}

Deep neural networks paved the way for significant improvements in image visual categorization during the last years. However, even though the tasks are highly varying, differing in complexity and difficulty, existing solutions mostly build on the same architectural decisions. This also applies to the selection of activation functions (AFs), where most approaches build on Rectified Linear Units (ReLUs). In this paper, however, we show that the choice of a proper AF has a significant impact on the classification accuracy, in particular, if fine, subtle details are of relevance. Therefore, we propose to model the degree of absence and the presence of features via the AF by using piece-wise linear functions, which we refer to as L*ReLU. In this way, we can ensure the required properties, while still inheriting the benefits in terms of computational efficiency from ReLUs. We demonstrate our approach for the task of Fine-grained Visual Categorization (FGVC), running experiments on seven different benchmark datasets. The results do not only demonstrate superior results but also that for different tasks, having different characteristics, different AFs are selected.

\end{abstract}

%% file: sections/introduction.tex
\section{Introduction}

Deep Convolutional Neural Networks (\eg, \cite{pr:goodfellow16,pr:lecun15} have recently shown to be very beneficial for a variety of applications in the field of Computer Vision. Thus, there has been, for instance, a considerable interest in designing new network architectures, introducing efficient data augmentation techniques, and improving the parameter optimization. In addition, to increase the robustness and the speed of training also different techniques for initialization (e.g., \cite{pr:sutskever13,pr:mishkin15}) and normalization (e.g., \cite{pr:laurent16}) have been explored. However, one relevant and important parameter is mostly ignored, the proper choice of the non-linear activation function (AF).

\begin{figure}[!t]
\centering
\begin{subfigure}{0.24\textwidth}
	\centering
	\includegraphics[width=\textwidth]{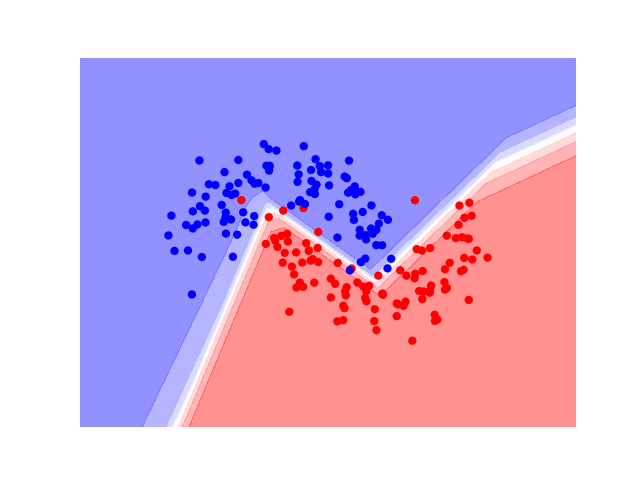}
	\caption{\emph{L*ReLU} (proposed).}
	\label{fig:toy:a}		
\end{subfigure}%
\begin{subfigure}{0.24\textwidth}
	\centering
	\includegraphics[width=\textwidth]{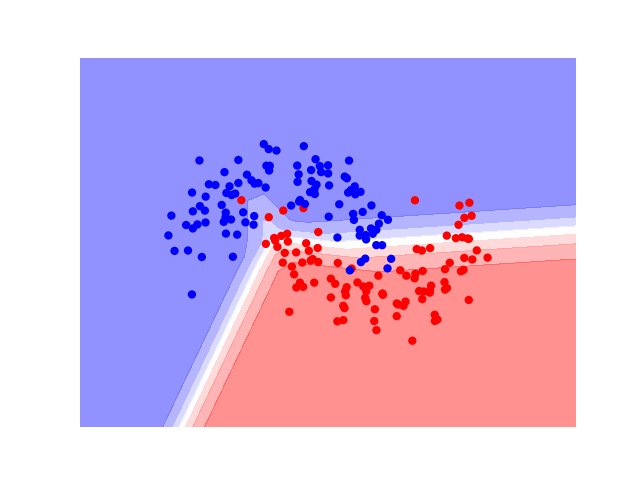}
	\caption{\relu.}
	\label{fig:toy:b}
\end{subfigure}%
\caption{Binary classification for the two-moon dataset: Just by using a proper activation function a better decision boundary can be estimated.} 
\label{fig:teaser}
\end{figure}

In fact, recent works have demonstrated that introducing \cite{pr:klambauer17,pr:clevert16,pr:elfwing18,pr:glorot11,pr:gulcehre16}) and learning \cite{pr:ramachandran18,pr:basirat19,pr:hayou18,pr:li18a} new AFs and their parameters are beneficial in terms of convergence speed and training stability, however, also that only minor improvements for the final tasks can be achieved. Thus, most deep learning approaches use Rectified Linear Units (\relu) \cite{pr:nair10}, which have proven to be reliable and allow for fast learning. In this paper, however, we show that the proper choice of the AF can significantly improve the performance of deep neural networks for applications like segmentation, tracking, or object retrieval where subtle visual differences are of relevance. In particular, we demonstrate these benefits for Fine-grained Visual Categorization (FGVC) \cite{mb:lin15a,mb:krause14}.

In contrast, to Coarse-grained Visual Categorization (CGVC), which aims at distinguishing well-defined categories (\eg, dogs, birds, or man-made objects), the goal of FGVC \cite{mb:lin15a,mb:krause14} is to differentiate between hard-to-distinguish classes (\eg, classes belonging to the same category such as different species of birds). Thus, there is a high visual similarity between the classes, where even subtle differences are of relevance. This is illustrated in Fig.~\ref{fig:teaser}, where we trained a shallow network (2 layers) for a two-class toy problem (\ie, the two-moon dataset) using different activation functions, namely \relu and the proposed \emph{L*ReLU} with $\alpha = 0.1$). In fact, using the same architecture a better decision boundary can be estimated. In particular, the samples close to the true decision boundary, having a small distance to each other, can be classified significantly better.

In this paper, we address the FGCV problem by using a proper AF, modeling the degrees of presence and absence of features \cite{pr:clevert16}. If a feature is present in the current sample, an AF returns a value greater than zero; on the other hand, if a feature is absent, a value smaller or equal to zero is returned. For example, \relu maps the presence of the features via identity, whereas all missing features are mapped to zero. For FGCV, however, it is important also to retain the degree of absence, \ie, the output of the AF in the deactivation state should not be zero.

To ensure the desired properties, we need an AF which is monotonically increasing and uniform-continuous. In other words, the negative values should not saturate, and similar inputs should produce similar outputs. We are ensuring these properties via a piece-wisely defined (positive and negative domain) linear AFs, where the positive part is the identity function and the negative part a linear function with a data-dependent slope. Thus, we refer our method to as \emph{L*ReLU}, indicating the similarities to \emph{Leaky ReLU} (having a slope of $0.01$) and the fact that the slope is set according to a data-dependent Lipschitz constant \cite{cohen2019universal}. In this way, not only the desired properties are ensured but also the positive properties of \relu are inherited. The experimental results on seven different datasets clearly show the benefits in terms of classification accuracy compared to the baselines, but also that for different tasks different AFs (\ie, different parameterizations) are necessary.

Thus, the main contributions of this paper can be summarized as follows:

\begin{itemize}

\item We propose to model the degree of absence and the presence of features via a properly defined activation function (AF).

\item We propose \emph{L*ReLU}, a piece-wise linear AF, where the slope of the negative part is selected according to a properly defined, data-dependent Lipschitz constant.

\item We run a thorough experimental analysis on five different Fine-grained Visual Categorization (FGCV) benchmarks and compared to both pre-defined as well as parametric AFs. The results clearly demonstrate the benefits of the approach in terms of improved classification accuracy and applicability for multiple tasks.

\end{itemize}


%% file: sections/related.tex
\section{Related Work}
\label{sec:related}

In the following, we first give a short review on Fine-grained Visual Categorization (FGVC), then discuss in detail different AFs for deep network training, and finally summarize the idea of Lipschitz regularization.

\subsection{Fine-grained Visual Categorization} 

To provide a sufficient discriminative capability for FGVC, several techniques have been explored over time. For instance, this can be achieved by learning discriminative features \cite{mb:sanchez11,mb:perronnin15}, where recently, in particular, Bilinear-CNNs, which compute second-order bilinear features interactions using two symmetric CNNs \cite{mb:lin15a,mb:gao15,mb:kong17,mb:cai17,mb:cui17} have shown to work very well in practice. Alternative approaches are focusing on local information and model parts of objects \cite{mb:krause14,mb:zhang14,mb:huang16,mb:shi15,mb:lin15b,mb:simon15,mb:krause15}. To identify and detect informative regions which include important local information, recently, attention models have been emerging \cite{mb:yu18, mb:sun18,mb:jaderberg15,mb:chen18,mb:fu17}. In addition, also metric learning approaches (\eg, \cite{mb:wang14}), which might be most similar to our approach, are applied. In contrast, in this work, we show that the FGVC problem can be addressed by applying more appropriate, task-specific AFs.

\subsection{Pre-defined Activation Functions}

Starting from simple thresholding functions, initially, the main focus when developing activation functions (AFs) was on squashing functions such as \func{Sigmoid} and \func{Tanh} \cite{pr:hornik91}. In particular, as following the universal approximation theorem \cite{pr:hornik91} any continuous real-valued function can be arbitrary well approximated by a feed-forward network with one hidden layer if the AF is continuous, bounded, and monotonically increasing. However, such functions are suffering from the vanishing gradient problem \cite{pr:hochreiter98}, which is, in particular, a problem if the networks are getting deeper. 

To overcome this problem, various non-squashing functions were introduced, where, in particular, \relu \cite{pr:nair10} paved the way for the success for deep learning. As the derivative of positive inputs of \relu is one, the gradient cannot vanish. On the other hand, all negative values are mapped to zero, resulting in two main problems: (1) There is no information flow for negative values, which is known as dying \relu. (2) The statistical mean of the activation values is still larger than zero, leading to a bias shift in successive layers. Moreover, all negative values are treated equally, which is not desirable for the FGVC task! To deal with the dying-\relu-problem \func{Leaky}~\relu (\lrelu) \cite{pr:maas13} introduces a very small negative slope ($\alpha=0.01$) for the negative part. Even though showing better results for many tasks, the function is still suffering from the bias shift. Slightly differently \func{Randomized Leaky Rectified Linear Unit} (\func{RReLU}) \cite{pr:xu15} sets the slope for the negative part randomly. 

Both shortcomings of \relu can be avoided by using \func{Exponential Linear Unit} (\elu) \cite{pr:clevert16}, which is robust to noise and eliminates the bias shift in the succeeding layers by pushing the mean activation value towards zero. By returning a bounded exponential value for negative inputs \func{ELU} is saturated at a predefined threshold. The idea was later extended by introducing \emph{Scaled Exponential Linear Unit} \selu \cite{pr:klambauer17}, showing that the proposed self-normalizing network converges towards a normal distribution with zero mean and unit variance. However, both \elu and \selu are bounded in the negative part, which is not a desired property for the FGVC task.

\subsection{Learned and Parametric Activation Functions}

To increase the flexibility, parametric AFs have been proposed, which learn parameters to tune themselves during the training. For instance, \func{Parametric} \relu (\prelu) \cite{pr:he15} builds on the ideas of \lrelu, but learns the slopes for the negative part based on the training data. Moreover, SReLU \cite{Srelu-AF} is defined via three piece-wise linear functions including four adaptive scalar values, forming a crude S shape. Having both convex and non-convex shapes are remarkable characteristics of SReLU.

Similarly, \func{Parametric}~\elu \cite{pr:trottier17} evades the vanishing gradient problem and allows for precisely controlling the bias shift by learning parameters from the data. More complex functions (\ie, even non-convex ones) can be learned using \func{Multiple Parametric Exponential Linear Units} \cite{pr:li18a}, which, in turn, leads to a better classification performance and preferable convergence properties.

The same goal can also be achieved by adopting ideas from reinforcement learning \cite{pr:ramachandran18} and genetic programming \cite{pr:basirat19}, where complex search spaces are explored to construct new AFs. In particular, in \cite{pr:ramachandran18} \swish, a combination of a squashing and a linear function, was found as the best solution for a variety of tasks. Moreover, \emph{Parametric} \swish (\pswish) contains a trainable (scaling) parameter. Similar functions, also yielding slightly better results in the same application domains, were also found by \cite{pr:basirat19}. Recently, a theoretic justification for these results have been given in \cite{pr:hayou18}, showing that \swish-like functions propagate information better than \relu. 

However, even though these functions yield good results for CGVC, they cannot cope very well with FGVC tasks. Thus, the goal of this paper is to define AFs better suited for more complex image classification tasks, capturing the small, subtle differences between the rather similar classes more effectively.

\subsection{Lipschitz Regularization}

Motivated by fact that even small perturbations on the input data can significantly change the output (``adversarial samples'') \cite{pr:Szegedy14, TowardsFast}, there has been considerable interest in Lipschitz regularization for deep neural network training. Recent works show that given a constrained Lipschitz constant is meaningful for DNNs in terms of robustness and classification accuracy \cite{lipchitzregualrity,fazlyab,cohen2019universal}. The Lipschitz constant bounds the ratio of output change to change in its input. In particular, for classification tasks, a small Lipschitz constant improves generalization ability \cite{largemargin,sorting}. In this way, in \cite{l2nonexpansive} a $L_2 $-nonexpansive neural network is introduced to control the Lipschitz constant and increase the robustness of classifier. Similarly, a robust deep learning model using Lipschitz margin is proposed for object recognition \cite{lipschitzmargin}. These works mainly intended to increase the robustness against adversarial samples to guarantee better convergence properties. In contrast, we adopt ideas that relate the separability of multiple classes to the choice of a proper Lipschitz constant for piece-wise linear AFs. This additionally ensures the desired properties to model the degrees of absence and presence of features very well.

%% file: sections/method.tex
\section{{L*ReLU}: Lipschitz ReLU}
\label{sec:nipcaf}

In the following, in Sec.~\ref{sec:pasf}, we first discuss the problem of modeling the presence and the absence of features by via AFs. Then, we discuss technical preliminaries on continuity of functions in Sec.~\ref{sec:cont-functions}. Finally, in Sec.~\ref{sec:opt-slope}, we introduce 
\emph{L*ReLU}, which copes with the similarity of samples in the FGVC problem much better.

\subsection{Presence and Absence of Features}
\label{sec:pasf}

The output $a_j$ of a single neuron $j$ within a neural network is computed by
\vspace{-0.25cm}
\begin{equation}
a_j = f\left(\sum_{i=0}^n w_{i,j} a_i\right) ,
\vspace{-0.25cm}
\end{equation}

\noindent where $a_i$ are the outputs of the $n$ connected neurons (from the previous layer), $w_{i,j}$ are the related weights, and $f(x)$ is a non-linear function referred to as an activation function (AF). In this way, $f(x)$ codes the degree of presence or absence of a feature in the input \cite{pr:clevert16}: a feature is present if $f(x) > 0$  and absent if $f(x) \leq 0$. 

The degree of presence is modeled very well for most existing AFs. This can be seen from Figure~\ref{fig:AFs}, where we show well-known and widely used AFs, which are defined in Table~\ref{table:AFs}. In these cases, this is achieved via the positive part of the function (\ie, for $x > 0$) being either a linear function (\relu, \lrelu, \elu ) or a "quasi-linear" function (\swish).

\input{plots/fig-AFs}
%
\input{tables/table-AFs}

On the other hand, the degree of absence of features is not captured very well. For instance, \relu does define a clear ``off-state'' or ``deactivation-state'': $f(x)=0$ for all negative values. In this way, the neuron does not model any information about the degree of absence which can be propagated to the next layer. Similar also applies to \elu. Even though $f(x)$ is getting smaller if $x$ is decreased, the function saturates at $-1$, which represents the ``off-state''. As a result, the derivations for small values are getting smaller, thus, reducing the information that is propagated to the next layer. \swish, in contrast, models the degree of absence for small negative values well but also saturates at $0$ if the values are further decreased.

Taking only the presence of features into account is sufficient for many applications including CGVC. If a feature is absent, it is sufficient to say ``It's not there!''. Therefore, AFs such as \relu, \elu, \selu, and \swish are well suited for CGVC tasks. This is also revealed by our experimental results, where we show that for CGVC the choice of the AF has only a minor impact on the finally obtained accuracy. However, just modeling the presence is not sufficient for FGVC, where the different classes often share a similar appearance and often differ just in subtle visual differences. In these cases, also modeling the degree of absence is necessary. To define functions showing the necessary properties, we need first to review uniform- and Lipschitz-continuous functions.

\subsection{Uniform- and Lipschitz-continuous Functions}
\label{sec:cont-functions}

Even though the following concepts are more general, for reasons of simplicity, we will restrict the discussion to functions in $\mathbb{R}$ \cite{pr:gordon95, pr:sagan74}. 

A function $f: \mathbb{R} \rightarrow \mathbb{R}$ is called \emph{Lipschitz-continuous} if there exist a constant $L \geq 0$ such that
\begin{align}
\label{eq:lipschitz}
    | f(x_i)-f(x_j) | \leq L | x_i - x_j |    
\end{align}

\noindent for all $x_i, x_j \in \mathbb{R}$. Any $L$ fulfilling  the condition Eq.~(\ref{eq:lipschitz}) is referred to as a \emph{Lipschitz constant}. The minimum $\hat{L}$ of all Lipschitz constants $L$ is often called the \emph{ minimal Lipschitz constant}.
For $x_i \neq x_j$ we can re-write Eq.~(\ref{eq:lipschitz}) to 
\begin{align}
\label{eq:lc2}
   \frac{| f(x_i)-f(x_j) |}{| x_i - x_j |} \leq L .
\end{align}
  
\noindent This means that the slopes of secants and tangents in an interval $I \in \mathbb{R}$ are bounded by $L$. In particular, we have 

$$
f'(z) \leq L \text{~~~~for all~~} z \in I, 
$$

\noindent or in other words: 
$$
L = \sup_{x \in I} |f'(z)|.
$$
  
In this way, the Lipschitz constant $L$ measures the maximum change rate of function $f$ within an interval $I$.  If $ 0 \leq L < 1$, then $f$ is called a \emph{contraction mapping} on $I$. 

Moreover, a Lipschitz-continuous function $f:\mathbb{R} \rightarrow \mathbb{R}$ is also \emph{uniformly continuous}, that is, for every $\epsilon >0$ there exits a $\delta >0$ such that for all $x_i, x_j \in \mathbb{R}$ we have
 \begin{align}
| x_i - x_j | < \delta \Rightarrow | f(x_i) - f(x_j) | < \epsilon .
 \end{align}
 
In other words, a uniform-continuous functions ensures that $f(x_i)$ and $f(x_j)$ are close to each other if $x_i$ and $x_j$ are sufficiently close to each other.

\subsection{Piece-wise Linear Activation Functions}
\label{sec:opt-slope}

From the discussion above, it is clear that for FGVC an AF is needed, which models both the degree of presence and the degree of absences of features. As these two aspects are related to positive and negative domain of $\mathbb{R}$, we propose to use a piece-wise function for the positive values $x > 0$ and the negative values $x \leq 0$:
\begin{equation}
f(x)= p(x> 0) + n(x \leq 0) ,
\end{equation}
\noindent with 
\begin{equation}
p(x) = \max(\phi(x),0)
\end{equation}
\noindent and
\begin{equation}
n(x) = \min(\eta(x),0) ,
\end{equation}

\noindent where $\phi(x)$ and $\eta(x)$ be any (non-linear) function $f: \mathbb{R} \rightarrow \mathbb{R}$. In this way, we ensure that the positive and the negative part of the piece-wise function reside in the first and third quadrants of a Cartesian coordinate system. It is easy to see that we can easily re-write almost all popular AFs to such a form.

Modeling the presence of features is already realized very well by existing AFs using linear (\relu, \elu) or quasi-linear (\swish) functions for the positive domain. Modeling the degree of absence of features, however, is more difficult. In particular, we would need a non-saturating, monotonically increasing function with bounded change rate (\ie, similar inputs should generate similar outputs). Given the definitions from Sec.~\ref{sec:cont-functions} this means that we need contractive, unbounded Lipschitz- and thus also uniform-continuous functions. 

However, as can be seen from the example shown in  Sec.~\ref{sec:example}, defining such AFs is not trivial. Thus, we propose to use a piece-wise linear approximation:

\begin{equation}
p(x)=\max(x,0)
\end{equation}
\noindent and
\begin{equation}
n(x)=\min(\alpha x,0) ,
\end{equation}

\noindent where $\alpha \geq 0$ defines the slope of the linear function for the negative part. 

Indeed, using the parameter setting $\alpha=0.01$, we get the well-known \lrelu activation function. However, as we also show in the experiments using such small slopes, which are typically working well for CGVC, fails for FGVC, as the  degree of absence of features cannot be modeled very well. The same also applies to \rrelu, where the slopes are chosen randomly. Thus, a valid choice would be to use \prelu, where the slope parameters are estimated from the data. However, the method is not flexible enough in practice, as the optimizer often gets stuck into local minima, which are not generalizing very well. Thus, an initialization close to the optimal solution is needed. We also demonstrate this behavior in the experimental results, where we show that using \prelu only gives reasonable results if the approach was initialized close to the optimal solution.

Thus, the critical question is, how to optimally set the slope $\alpha$? Following the ideas of \cite{pr:basirat19} and \cite{cohen2019universal}, we argue that there is no unique solution across different tasks. In particular, \cite{pr:basirat19} shows that for classification tasks of different complexity different AFs are useful. In contrast, \cite{cohen2019universal} proves that given any finite dataset where different classes are separated in the input space by at least a distance of $c$, there exists a function with Lipschitz constant $c/2$ that correctly classifies all points. In other words, different datasets might have different separability, raising the need for learning functions with different properties. This can be realized by choosing a proper slope according to the Lipschitz-properties of the data. 

That is, we call our approach \emph{L*ReLU}, indicating both that the used AF builds on the ideas of \emph{Leaky ReLU} (and PReLU), but also that the slope parameter is chosen according to the Lipschitz-properties of the data. Indeed, our experiments also demonstrate that for the seven different datasets different parameters are needed, however, also that these are not critical. In practice, or each task we can identify a restricted range for the Lipschitz parameter (and thus the slope) \cite{fazlyab}, yielding stable and reliable classification results. 


%% file: plots/fig-AFs.tex
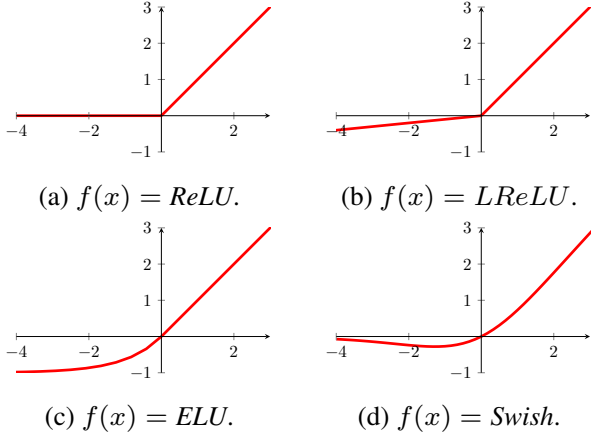
\begin{figure}[ht!]
\centering
\resizebox{0.5 \textwidth}{!}
{
\begin{tabular}{cc}
    \input{plots/ReLu.tex} &
    \input{plots/LReLu.tex} \\
    {\footnotesize (a) $f(x) = \relu$.} &
    {\footnotesize (b) $f(x) = LReLU$.} \\
    \input{plots/ELU.tex} &
    \input{plots/swish.tex}  \\
    {\footnotesize (c) $f(x)=\elu$.} &
    {\footnotesize (d) $f(x) = \swish$.} \\
  \end{tabular}
}
\caption{Activation functions as defined in Table~\ref{table:AFs}.}
\label{fig:AFs}
\end{figure}

%% file: plots/ReLu.tex
\begin{tikzpicture}[scale=1/2]
    \begin{axis}
        [xmin=-4, xmax=3,
        ymin=-1, ymax=3,
        x=0.75cm,
        y=0.75cm,
        axis lines=center,
        axis on top=true]
        \addplot [mark=none,draw=red,ultra thick,domain=0:4] {\x};
        \addplot [mark=none,draw=red,ultra thick,domain=-10:0] {0};
\end{axis}
\end{tikzpicture}

%% file: plots/LReLu.tex
\begin{tikzpicture}[scale=1/2]
    \begin{axis}
        [xmin=-4, xmax=3,
        ymin=-1, ymax=3,
        x=0.75cm,
        y=0.75cm,
        axis lines=center,
        axis on top=true]
        \addplot [mark=none,draw=red,ultra thick,domain=0:4] {\x};
        \addplot [mark=none,draw=red,ultra thick,domain=-10:0] {0.1*\x};
\end{axis}
\end{tikzpicture}

%% file: plots/ELU.tex
\begin{tikzpicture}[scale=1/2]
\begin{axis}
    [xmin=-4, xmax=3,
    ymin=-1, ymax=3,
    x=0.75cm,
    y=0.75cm,
    axis lines=center,
    axis on top=true]

    \addplot [mark=none,draw=red,ultra thick,domain=0:4] {\x};
    \addplot [mark=none,draw=red,ultra thick,domain=-10:0] {exp(\x)-1};
    
    
\end{axis}
\end{tikzpicture}

%% file: plots/swish.tex
\begin{tikzpicture}[scale=1/2]
\begin{axis}
    [xmin=-4, xmax=3,
    ymin=-1, ymax=3,
    x=0.75cm,
    y=0.75cm,
    axis lines=center,
    axis on top=true]

    \addplot [mark=none,draw=red,ultra thick,domain=0:4] {x*1/(1+exp(-x))};
    \addplot [mark=none,draw=red,ultra thick,domain=-4:0] {x*1/(1+exp(-x))};
\end{axis}
\end{tikzpicture}

%% file: tables/table-AFs.tex


\begin{table}[!ht]
\renewcommand{\arraystretch}{1.45}
\centering
\resizebox{\columnwidth}{!}{%
\begin{tabular}{rp{1.2cm}l}
\toprule
\multicolumn{2}{c}{Name}
&
\multicolumn{1}{c}{Function}
\\
\midrule
\midrule
(a)
&
\relu 
&
$y(x)=\max(x,0)$
\\
\rule{0pt}{2ex} 
(b)
&
\func{LReLU}
&
$
y(x) =
\max(x,0)+\min(0.01 x,0)
$
\\
\renewcommand{\arraystretch}{1.4}
(c)
&
\elu
&
$
y(x) = \max(x,0)+\min(e^{x}-1,0)
$
\\
(d)
&
\swish 
&
$y(x) =x \cdot \text{sigmoid}(\beta x)$
\\
\bottomrule
\end{tabular}
}
\caption{Sample activation functions. For \swish the parameter $\beta=1$ is fixed, however, it can be trained via \pswish.}
\label{table:AFs}
\end{table}

%% file: sections/experiments.tex
\section{Experimental Results}
\label{sec:experiments}

To demonstrate the importance and the effect of using proper AFs, we run experiments on seven different benchmark datasets: (1) two coarse-grained datasets (\ie, CIFAR-10 and CIFAR-100) \cite{pr:krishevski09} and (2) five fine-grained (\ie, Caltech-UCSD Birds-200-2011 \cite{mb:wah11}, Car Stanford \cite{mb:krause13}, Dog Stanford \cite{mb:khosla11}, Aircrafts \cite{mb:maji13}, and iFood \cite{mb:iFood}). We compared our approach to existing AFs, known to yield good results in practice, namely \relu \cite{pr:nair10}, \elu \cite{pr:clevert16}, \selu \cite{pr:klambauer17}, and \swish \cite{pr:ramachandran18}. Moreover, we also compare to parametric AFs, namely \prelu \cite{pr:trottier17} and \pswish \cite{pr:ramachandran18}, showing that \emph{L*ReLU} with a proper selected Lipschitz constant yields better results compared to all of these baselines.

In the following, we first describe the used benchmarks and the  experimental setup and then discuss the results for both the coarse-grained and the fine-grained datasets in detail.

\subsection{Benchmark Datasets}

\begin{figure}[!ht]
\centering
\begin{subfigure}{0.22\textwidth}
	\centering
	\includegraphics[width=\textwidth,height=50.8pt]{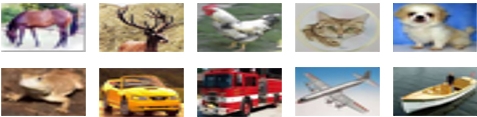}
	\caption{CIFAR-10.}
	\label{fig:datasets:cifar}		
\end{subfigure}
\hfill
\begin{subfigure}{0.22\textwidth}
	\centering
	\includegraphics[width=\textwidth]{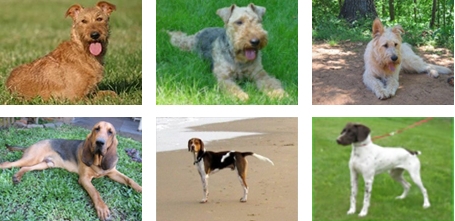}
	\caption{Dog Stanford.}
	\label{fig:datasets:dog}		
\end{subfigure}
\hfill
\begin{subfigure}{0.22\textwidth}
	\centering
	\includegraphics[width=\textwidth]{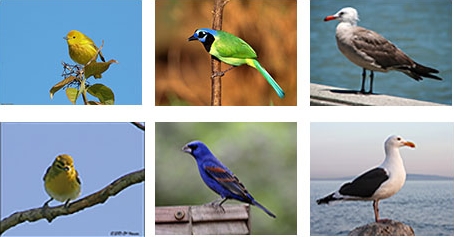}
	\caption{Caltech Bird-UCSD.\vspace{0.25cm}}
	\label{fig:datasets:bird}
\end{subfigure}
\hfill
\begin{subfigure}{0.22\textwidth}
	\centering
	\includegraphics[width=\textwidth]{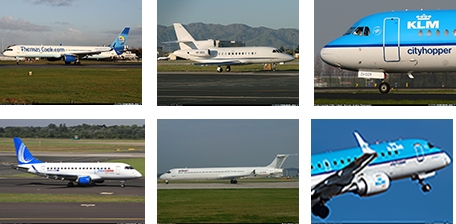}
	\caption{Aircraft.}
	\label{fig:datasets:aircraft}
\end{subfigure}
\hfill
\begin{subfigure}{0.22\textwidth}
	\centering
	\includegraphics[width=\textwidth]{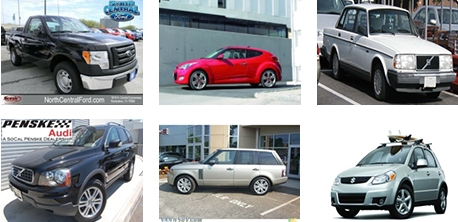}
	\caption{Car Stanford.}
	\label{fig:datasets:car}
\end{subfigure}
\hfill
\begin{subfigure}{0.22\textwidth}
	\centering
	\includegraphics[width=\textwidth]{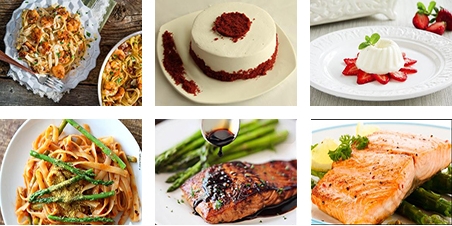}
	\caption{
	iFood.}
	\label{fig:datasets:food}
\end{subfigure}
\caption{Fine-grained visual categorization benchmark datasets (b)--(f) plus CIFAR-10 (a) used in our studies.}
\label{fig:datasets}
\end{figure}

The benchmark datasets described below are illustrated in Fig.~\ref{fig:datasets}. It can be seen that for the coarse-grained problem (Fig.~\ref{fig:datasets}~(a)) the single classes are well defined, whereas for the fine-grained problem (Figs.~\ref{fig:datasets}~(b)--(f)), the differences are just subtle and often hard to see. 
 
\paragraph{CIFAR-10 and CIFAR-100}
The CIFAR-10 dataset consists of 60,000 $32 \times 32$ colour images in 10 classes, with 6,000 images per class. CIFAR-100 is just like the CIFAR-10, except that it has 100 classes containing 600 images. The data is split into 5,0000 training images and 10000 test images.

\paragraph{Caltech-UCSD Birds-200-2011}
The dataset contains 11,788 images of 200 bird species. Each species is associated with a Wikipedia article and organized by scientific classification (order, family, genus, species). The data is split into 5,994 training images and 5,794 test images \cite{mb:wah11}).

\paragraph{Car Stanford}
The dataset contains 16,185 images of 196 classes of cars. The data is split into 8,144 training images and 8,041 test images, where each class has been split roughly in a 50-50 split. Classes are typically at the level of Make, Model, Year, e.g. 2012 Tesla Model S or 2012 BMW M3 coupe \cite{mb:krause13}.

\paragraph{Dog Stanford}
The dataset contains images of 120 breeds of dogs from around the world. This dataset has been built using images and annotation from ImageNet for the task of fine-grained image categorization. The total number of images is 20,580. The data is split into 12000 training images and 8,580 test images \cite{mb:khosla11}.

\paragraph{Air crafts}
The dataset contains 9960 images of aircraft, with 100 images for each of the 102 different aircraft model variants \cite{mb:maji13}.  The data is split into 3216 training, 3231 test and 3231 validation images.

\paragraph{iFood Dataset}
This large data-set consist of 211 fine-grained (prepared) food categories with 101733 training images collected from the web. Test set contains 10323 images. images\footnote{https://sites.google.com/view/fgvc5/
competitions/fgvcx/ifood}.

\subsection{Experimental Setup}
\label{sec:exp-setup}

To allow for a fair comparison, for all experiments the same experimental setup was used. In particular, to keep the computational cots at a reasonable level (\ie, one NVIDIA Titan XP GPU was used), we re-sized all images to a size of $120 \times 120$.  We trained an architecture similar to \emph{VGG} cosisting of eight convolutional layers plus two fully connected layers with $400$ and $900$ units, respectively. Moreover, we used a batch size of $70$ and set the maximum pooling size to seven. For training an Adam optimizer with batch normalization was applied. Since we used a different image size and the weights are related to the used AF, we were not able to use pre-trained weights. Thus, we used a random initialization for each training, but---to ensure statistically fair results---we run all experiments three times, where the mean results (and the standard deviations) are shown, respectively.



\subsection{Coarse-grained Visual Categorization}

\begin{figure}[!ht]
\centering
\begin{subfigure}{0.20\textwidth}
	\centering
	\includegraphics[page=6, width=\textwidth]{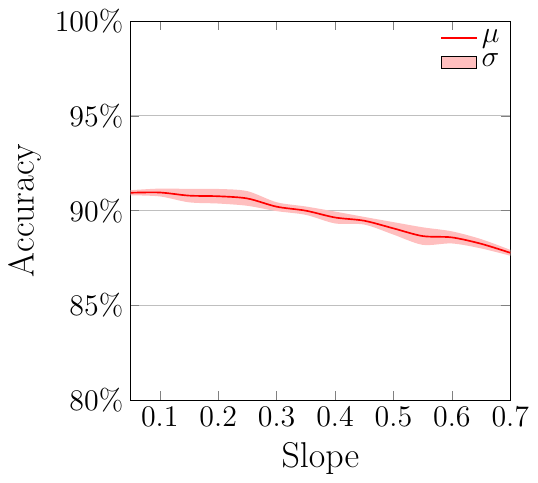}
	\caption{CIFAR-10}
	\label{fig:cslope:a}		
\end{subfigure}%
\begin{subfigure}{0.20\textwidth}
	\centering
	\includegraphics[page=7, width=\textwidth]{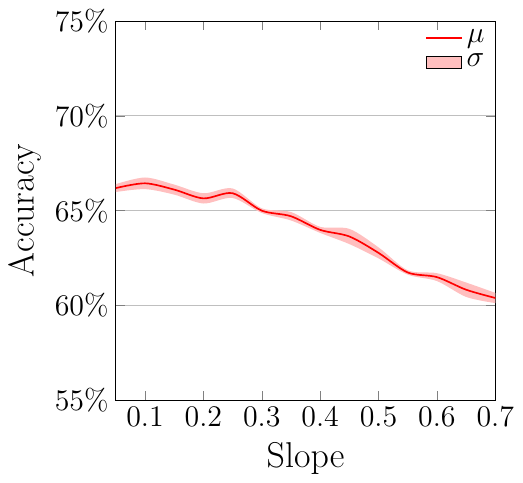}
	\caption{CIFAR-100}
	\label{fig:cslope:b}
\end{subfigure}%
\caption{Class. accuracy of \emph{L*ReLU} vs. slope for CGVC.}
\label{fig:cifar-slopes}
\end{figure}

\begin{figure*}[!ht]
\centering
	\begin{subfigure}{0.35\textwidth}
	\centering
	\includegraphics[page=9, width=\textwidth]{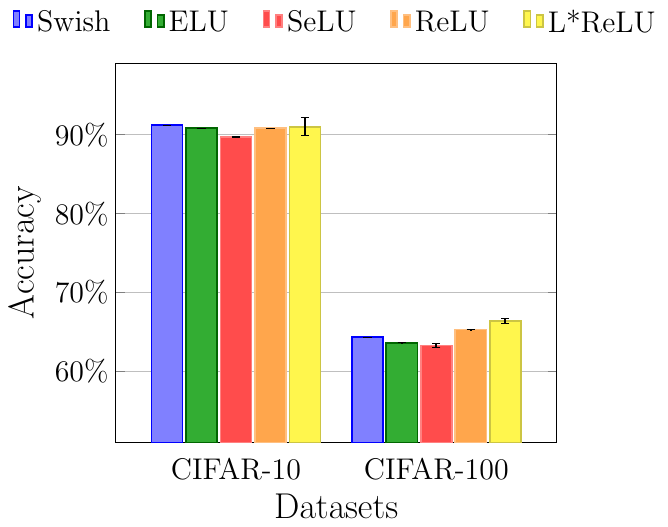}
	\caption{}
	\end{subfigure}%
\begin{subfigure}{0.64\textwidth}
	\centering
	\includegraphics[page=8, width=\textwidth]{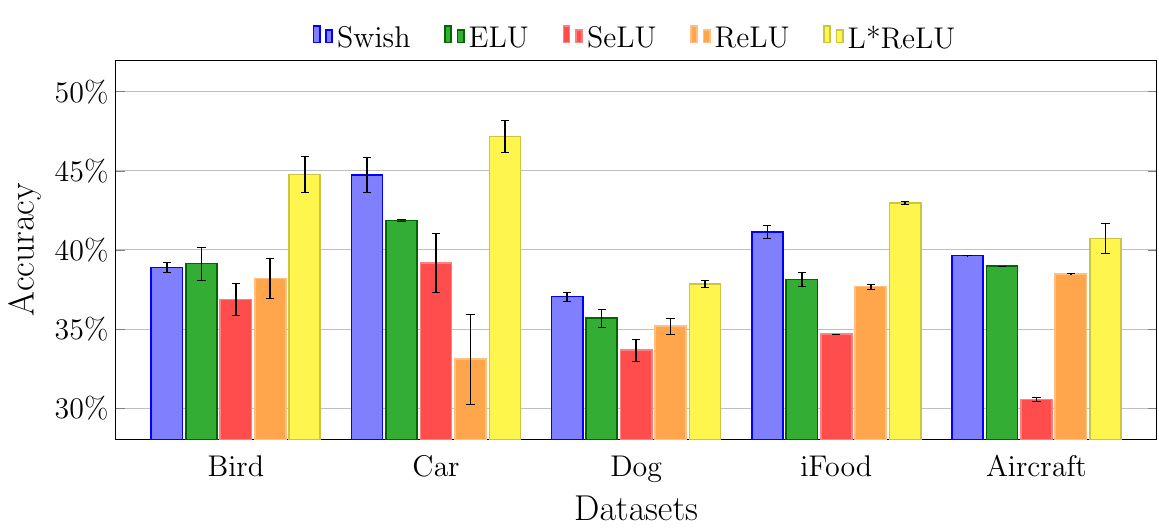}
	\caption{}
	\end{subfigure}%

	\caption{Mean and standard deviation of accuracy for all datasets and AFs: (a) CGVC and (b) FGVC.}
\label{fig:fg:accuracy_all}
\end{figure*}

First of all, we evaluated our approach for the coarse-grained datasets (\ie, CIFAR-10 and CIFAR-100). In Table~\ref{table:results-cifar} we show the final averaged results for all methods. The same results also covering the standard deviation are shown in form of boxplots in Fig.~\ref{fig:fg:accuracy_all}~(a). In addition, in Figure~\ref{fig:cifar-slopes}, we analyze the classification accuracy when varying the slope. Also here, in addition to the mean, the standard deviation is shown. For CIFAR-10 it can be seen that all AFs perform on par, where \swish slightly outperforms the others (which confirms previously published studies). In addition, as can be seen from  Figure~\ref{fig:cifar-slopes}~(a), increasing the slope for \emph{L*ReLU} decreases the classification accuracy, showing that the data is already well separable and that further enforcing the separability is not helpful. 

In contrast, for CIFAR-100 we can observe slightly different results. Figure~\ref{fig:cifar-slopes}~(b) also shows a clear trend that increasing the slope decreases the classification accuracy of \emph{L*ReLU}. However, a slight slope in the range of $0.1$ to $0.25$ (with a peak at $0.1$) demonstrated to be beneficial (compared to \relu) and allows to outperform the baselines. Due to the higher number of classes (\ie, $100$ instead of $10$), the task is more complex as the classes are getting more similar. Thus, modeling the degree of absence is getting relevant.

\begin{table}[!ht]
\centering
\resizebox{\columnwidth}{!}{%
\begin{tabular}{lccccc}
\toprule
Dataset & ELU & Swish & ReLU & SeLU&\textbf{L*ReLU}\\
\midrule
\midrule
CIFAR-10 & 90.81\% & \textbf{91.23}\% & 90.83\% &89.72\%& \emph{90.95\%} \\
\midrule
CIFAR-100 & 63.64\% &64.36\% & \emph{65.32\%} &63.29\% &\textbf{66.44\%} \\
\bottomrule
\end{tabular}
}
\caption{Mean accuracy for CGVC: The best result is in \textbf{boldface}, the runner up in \emph{italic}.}
\label{table:results-cifar}
\end{table}

\begin{figure*}[!ht]
\centering
\begin{subfigure}{0.20\textwidth}
	\centering
	\includegraphics[page=1, width=\textwidth]{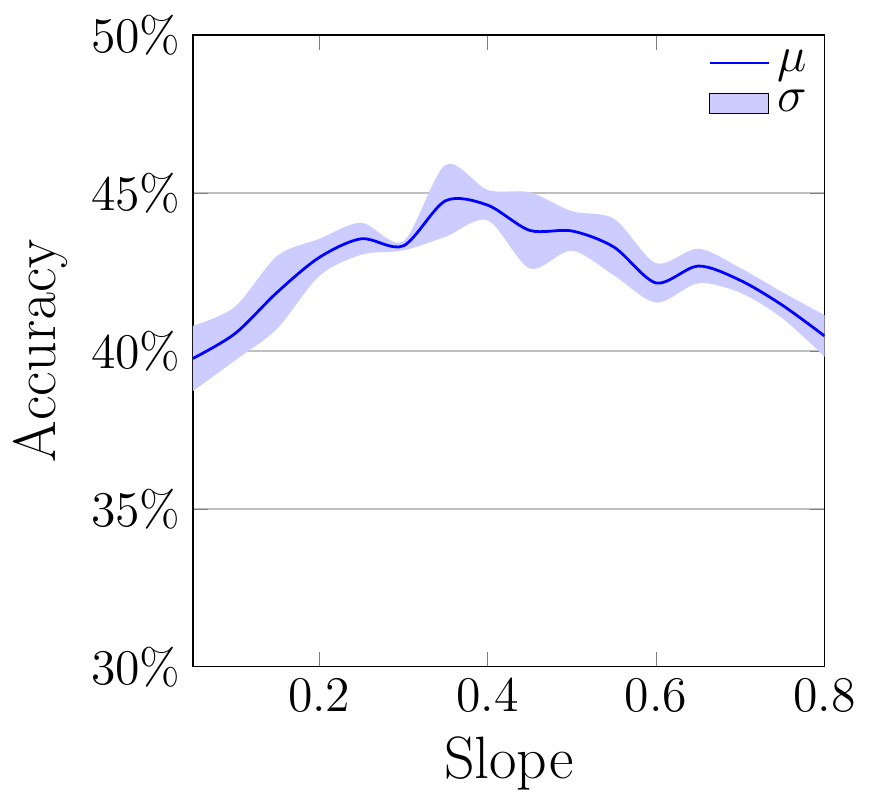}
	\caption{Bird.}
	\label{fig:slope:a}		
\end{subfigure}%
\begin{subfigure}{0.20\textwidth}
	\centering
	\includegraphics[page=2, width=\textwidth]{results/Tizit_plot_BMVC_withouttitle1.pdf}
	\caption{Car.}
	\label{fig:slope:b}
\end{subfigure}%
\begin{subfigure}{0.20\textwidth}
	\centering
     \includegraphics[page=3, width=\textwidth]{results/Tizit_plot_BMVC_withouttitle1.pdf}
	\caption{Dog.}
	\label{fig:slope:c}
\end{subfigure}%
\begin{subfigure}{0.20\textwidth}
	\centering
	 \includegraphics[page=4, width=\textwidth]{results/Tizit_plot_BMVC_withouttitle1.pdf}
	\caption{iFood.}
	\label{fig:slope:d}
\end{subfigure}%
\begin{subfigure}{0.20\textwidth}
	\centering
	 \includegraphics[page=5, width=\textwidth]{results/Tizit_plot_BMVC_withouttitle1.pdf}
	\caption{Aircraft.}
	\label{fig:slope:e}
\end{subfigure}
\caption{Class. accuracy of \emph{L*ReLU} vs. slope for FGVC.}
\label{fig:finegrained-slopes}
\end{figure*}

\subsection{Fine-grained Visual Categorization}

Next, we run the same experiments for the more complex FGVC task. 
\comment{Compared to the CGVC datasets the number of classes and the number of images is larger. In addition,}
As discussed above, the single classes are more similar as they represent the same general category, making it harder to distinguish the single instances and classes. The finally obtained averaged classification accuracy for all five datasets and all AFs are summarized in Table~\ref{table:results-finegrained}. The same results also covering the standard deviation are shown in form of boxplots in Fig.~\ref{fig:fg:accuracy_all}~(b). The results show that for all datasets \emph{L*ReLU} finally yields the best and \swish the second best results. Whereas the results are close ($+1\%$) for \emph{Dogs} and \emph{Aircraft} for the best and the second best results, the gap is larger for \emph{iFood}, \emph{Car}, and \emph{Birds}: up to $+5\%$. Moreover, it is notable that compared to \relu, which can be seen as a baseline, there is a significant gap for all datasets.

\begin{figure*}[!ht]
\centering
	\begin{subfigure}{0.5\textwidth}
	\centering
	\includegraphics[page=5, width=\textwidth]{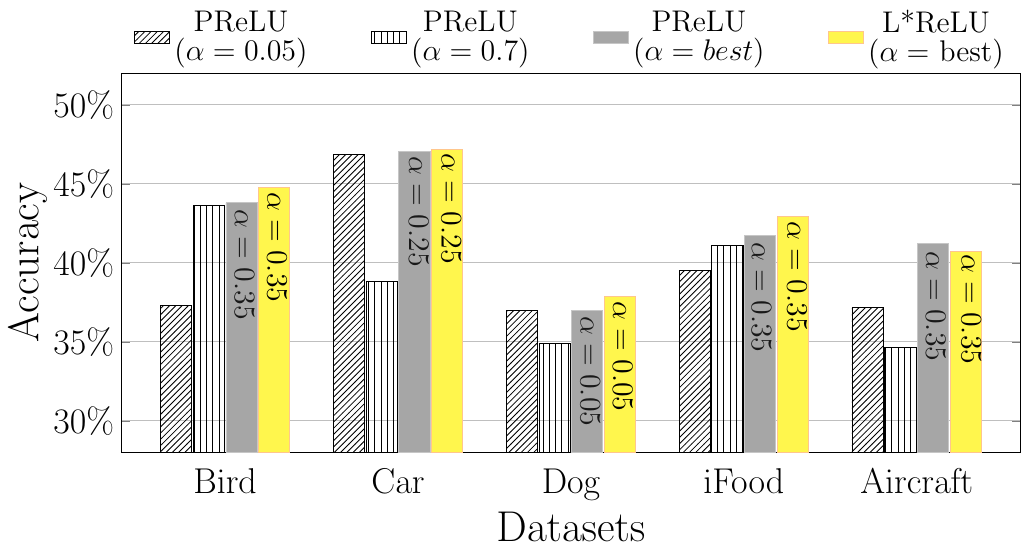}
	\caption{}
	\end{subfigure}%
	\hfill
		\begin{subfigure}{0.5\textwidth}
	\centering
	\includegraphics[page=6, width=\textwidth]{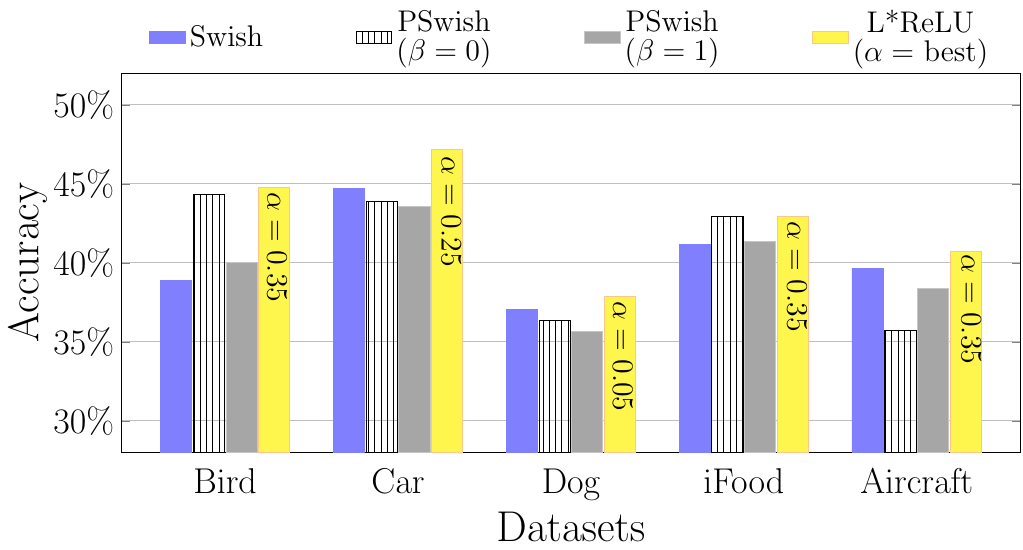}
	\caption{}
	\end{subfigure}%
\caption{Accuracy for  FGVC: \emph{L*ReLU} vs. parametric AFs: (a) \prelu and (b) \pswish.}
\label{fig:fg:prelu-comparision}
\end{figure*}

\begin{table}[!ht]
\centering
\resizebox{\columnwidth}{!}{%
 \begin{tabular}{lccccr}\toprule
 Dataset & ELU &  Swish & ReLU & SeLU &\textbf{L*ReLU}  \\
 \midrule
 \midrule
 Birds 200 & \emph{39.12\%} & 38.89\% & 38.18\% &36.86\% & \textbf{44.75}\%\\
 \midrule
  Car & 41.86\% & \emph{44.72}\% & 33.08\% & 39.17 \% &\textbf{47.16}\%\\
 \midrule
 Dogs & 35.67\% & \emph{37.04\%\emph} &  35.17\% &33.65\%& \textbf{37.85}\%\\
 \midrule
  iFood & 38.12\%&  \emph{41.14\%} & 37.67\% &34.67\%&\textbf{42.94}\%\\
  \midrule
  Aircraft & 38.97\%&  \emph{39.63\%} & 38.49\% &30.54\% & \textbf{40.72}\%\\
 \bottomrule
 \end{tabular}}
\caption{Mean accuracy for FGVC: The best result is in \textbf{boldface}, the runner up in \textit{italic}.}
\label{table:results-finegrained}
\end{table}

In addition, again we analyze the averaged classification accuracy (plus standard deviation) varying the slops in Fig.~\ref{fig:finegrained-slopes}. It can be seen that the different datasets define a different Lipschitz level, showing that applying task-specific AFs is meaningful. Moreover, also these curves show clear trends with clear peaks. Indeed, for all datasets there is constrained slope range of $[0.1, 0.4]$, where stable classification results are obtained. Thus, the selection of the Lipschitz constant for \emph{L*ReLU}is important, but not critical. 
%

\subsection{Comparison to Parametric AFs}

Next, we give a detailed comparison of \emph{L*ReLU} to parametric AFs, which should better adapt to more complex problems due to trainable parameters? In particular, we compare our approach to  \emph{Parametric ReLU} (\prelu) \cite{pr:trottier17} and \emph{Parametric Swish}  (\emph{PSwish}) \cite{pr:ramachandran18} using different initializations: for \prelu we used a small ($0.05$), a large (0.7), and the best slope for \emph{L*ReLU} to initialize $\alpha$; similarly, for \emph{PSwish} a small ($0.0)$ and large ($1.0$) value was used to initialize $\beta$ (see Table~\ref{table:AFs}). The corresponding results are shown in Fig.~\ref{fig:fg:prelu-comparision}, respectively. From Fig.~\ref{fig:fg:prelu-comparision}(a) it can be seen that the classification accuracy is highly varying depending on the initialization. Similar also applies for \emph{PSwish}, even though the variation in the accuracy is smaller and thus less sensitive to the initialization. 



\subsection{Importance of Lipschitz Constant}
 \label{sec:example}

\begin{figure}[ht!]
\centering
  \input{plots/tanhx+x.tex}
  \caption{Two functions sharing the same Lipschitz constant $L=0.25$ for the negative domain: $f(x) =tanh(0.1 x)\! + \! 0.15 x $ (red) and $g(x) = 0.25 x$ (blue).}
  \label{fig:tanhx}
\end{figure}
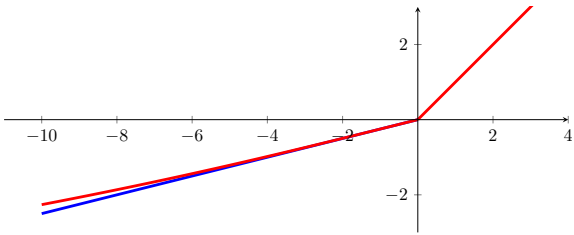

Finally, we would like to demonstrate the importance of the proper selected Lipschitz constant,
by comparing the accuracy of a more complex activation function \cite{pr:hayou18},
\begin{align}
f(x) = tanh(a x) + b x ,
\label{eq:exf1}
\end{align}
\noindent with our linear approximation 
\begin{align}
g(x) = \alpha x 
\label{eq:exg1}
\end{align}

\noindent based on the Lipschitz constant for the negative domain. For the positive domain, we used the identity function, respectively. However, in the following, we are only focusing on the negative domain! Both functions are illustrated in Fig.~\ref{fig:tanhx}.
From Eq.~(\ref{eq:exf1}) we compute the derivation
\begin{align}
f'(x) = \frac{1}{cos^2(a x)} + b . 
\label{eq:exf}
\end{align}
\noindent In this way, for $a = 0.1$ and $b=0.15$ we compute the Lipschitz constant via $\sup_{x \leq 0}(f'(x)) = 0.25$. Similarly, setting $\alpha=0.25$ the Lipschitz constant of $g(x)$ is $0.25$, as $g(x)$ is a linear function having the same slope for all $x \leq 0$. 

In this way, we set the parameters $a$ and $b$ of $f_{a,b}(x)$ such that the Lipschitz constant matches the best and the worst slope \emph{L*ReLU} for each datasets, respectively. The thus obtained results are shown in Table~\ref{table:tanh-vs-prelu}, showing that for all datasets we get a similar classification accuracy, indicating that the Lipschitz constant covers essential information about the data.

\begin{table}[!ht]
\centering
\resizebox{\columnwidth}{!}{
 \begin{tabular}{lcc|cr}\toprule
 Dataset & 
 \shortstack{\emph{L*ReLU}\\(best $\alpha$)} &
 \shortstack{$f_{a, b}(x)$\\(best $\alpha$)} & 
 {\shortstack{\emph{L*ReLU}\\($\alpha\!\!=\!\! 0.7$)}} &
 {\shortstack{$f_{0.4, 0.3}(x)$\\($\alpha\!\!=\!\! 0.7$)}}
 \\
 \midrule
 \midrule
 
  Car  & 47.16\% & 46.49\% (a)  & 33.31\% & 36.89\% \\
 \midrule
 Dogs & 37.85\% & 36.95\% (b) & 30.32\% & 31.06\% \\
 \midrule
 Birds 200  & 44.75\% & 43.16\% (c) & 42.23\% & 43.17\% \\
 \midrule
  iFood  & 42.94\% & 43.21\% (c) & 33.49\% & 35.24\% \\
  \midrule
  Aircraft & 40.72\% & 40.08\% (c) & 32.95\% & 35.04\% \\
 \bottomrule
 \end{tabular}
} 
\caption{Activation functions sharing the same Lipschitz constant finally yield a similar classification result: (a) $a\!\!=\!\!0.1$, $b\!\!=\!\!0.15$; (b) $a\!\!=\!\!0.05$, $b\!\!=\!\!0.05$; (c) $a\!\!=\!\!0.15$, $b\!\!=\!\!0.2$.}
\label{table:tanh-vs-prelu}
\end{table}

%% file: plots/tanhx+x.tex
\begin{tikzpicture}[scale=2/3]
\begin{axis}
    [xmin=-11, xmax=4,
    ymin=-3, ymax=3,
    x=0.75cm,
    y=0.75cm,
    axis lines=center,
    axis on top=true]
    \addplot [mark=none,draw=blue,ultra thick,domain=-10:0] {0.25*\x};
    \addplot [mark=none,draw=red,ultra thick,domain=0:4] {\x};
    \addplot [mark=none,draw=red,ultra thick,domain=-10:0] {tanh(0.1* \x) + 0.15* \x};
\end{axis}
\end{tikzpicture}

%% file: sections/conclusion.tex
\section{Conclusion and Discussion}
\label{sec:conclusion}


In this paper, we demonstrated that using a proper activation function can significantly improve the classification accuracy for the problem of Fine-grained Visual Categorization (FGVC), where subtle differences between similar images are of relevance. Thus, we propose to use activation functions, which model the degrees of presence and absence of features. Whereas the degree of presence is realized via an identity function, the degree of absence can be modeled via monotonically increasing uniform-continuous functions. In our case, we realized this by using piece-wise linear functions, where the slope of the negative part is set according to an optimal Lipschitz constant (give by the data). In this way, we outperform a wide range of fixed and parametric AFs for different FGVC benchmark datasets. Future work would include to automatically estimate the Lipschitz constant from the data and to explore the found properties for different, more complex AFs. 

%% file: main_review.bbl
\begin{thebibliography}{59}
\providecommand{\natexlab}[1]{#1}
\providecommand{\url}[1]{#1}
\csname url@samestyle\endcsname
\providecommand{\newblock}{\relax}
\providecommand{\bibinfo}[2]{#2}
\providecommand{\BIBentrySTDinterwordspacing}{\spaceskip=0pt\relax}
\providecommand{\BIBentryALTinterwordstretchfactor}{4}
\providecommand{\BIBentryALTinterwordspacing}{\spaceskip=\fontdimen2\font plus
\BIBentryALTinterwordstretchfactor\fontdimen3\font minus
  \fontdimen4\font\relax}
\providecommand{\BIBforeignlanguage}[2]{{%
\expandafter\ifx\csname l@#1\endcsname\relax
\typeout{** WARNING: IEEEtranN.bst: No hyphenation pattern has been}%
\typeout{** loaded for the language `#1'. Using the pattern for}%
\typeout{** the default language instead.}%
\else
\language=\csname l@#1\endcsname
\fi
#2}}
\providecommand{\BIBdecl}{\relax}
\BIBdecl

\bibitem[Goodfellow et~al.(2016)Goodfellow, Bengio, and
  Courville]{pr:goodfellow16}
I.~Goodfellow, Y.~Bengio, and A.~Courville, \emph{Deep Learning}.\hskip 1em
  plus 0.5em minus 0.4em\relax MIT Press, 2016.

\bibitem[LeCun et~al.(2015)LeCun, Bengio, and Hinton]{pr:lecun15}
Y.~LeCun, Y.~Bengio, and G.~Hinton, ``Deep learning,'' \emph{Nature}, vol. 521,
  pp. 436--444, 2015.

\bibitem[Sutskever et~al.(2013)Sutskever, Martens, Dahl, and
  Hinton]{pr:sutskever13}
I.~Sutskever, J.~Martens, G.~Dahl, and G.~Hinton, ``On the importance of
  initialization and momentum in deep learning,'' in \emph{Proc. Int'l Conf. on
  Machine Learning}, 2013.

\bibitem[Mishkin and Matas(2017)]{pr:mishkin15}
D.~Mishkin and J.~Matas, ``All you need is a good init,'' in \emph{Proc. Int'l
  Conf. on Learning Representations}, 2017.

\bibitem[Laurent et~al.(2016)Laurent, Pereyra, Brakel, Zhang, and
  Bengio]{pr:laurent16}
C.~Laurent, G.~Pereyra, P.~Brakel, Y.~Zhang, and Y.~Bengio, ``Batch normalized
  recurrent neural networks,'' in \emph{Proc. Int'l Conf. on Acoustics, Speech
  and Signal Processing}, 2016.

\bibitem[Klambauer et~al.(2017)Klambauer, Unterthiner, Mayr, and
  Hochreiter]{pr:klambauer17}
G.~Klambauer, T.~Unterthiner, A.~Mayr, and S.~Hochreiter, ``Self-normalizing
  neural networks,'' in \emph{Advances on Neural Information Processing
  Systems}, 2017.

\bibitem[Clevert et~al.(2016)Clevert, Unterthiner, and
  Hochreiter]{pr:clevert16}
D.~Clevert, T.~Unterthiner, and S.~Hochreiter, ``Fast and accurate deep network
  learning by exponential linear units ({ELUs}),'' in \emph{Proc. Int'l Conf.
  on Learning Representations}, 2016.

\bibitem[Elfwing et~al.(2018)Elfwing, Uchibe, and Doya]{pr:elfwing18}
S.~Elfwing, E.~Uchibe, and K.~Doya, ``Sigmoid-weighted linear units for neural
  network function approximation in reinforcement learning,'' \emph{Neural
  Networks}, vol. 107, pp. 3--11, 2018.

\bibitem[Glorot et~al.(2011)Glorot, Bordes, and Bengio]{pr:glorot11}
X.~Glorot, A.~Bordes, and Y.~Bengio, ``Deep sparse rectifier neural networks,''
  in \emph{Proc. Int'l Conf. on Artificial Intelligence and Statistics}, 2011.

\bibitem[Gulcehre et~al.(2016)Gulcehre, Moczulski, Denil, and
  Bengio]{pr:gulcehre16}
C.~Gulcehre, M.~Moczulski, M.~Denil, and Y.~Bengio, ``Noisy activation
  functions,'' in \emph{Proc. Int'l Conf. on Machine Learning}, 2016.

\bibitem[Ramachandran et~al.(2018)Ramachandran, Zoph, and
  Le]{pr:ramachandran18}
P.~Ramachandran, B.~Zoph, and Q.~V. Le, ``Searching for activation functions,''
  in \emph{Proc. Int'l Conf. on Learning Representations (Workshop track)},
  2018.

\bibitem[Basirat and Roth(2019)]{pr:basirat19}
M.~Basirat and P.~M. Roth, ``Learning task-specific activation functions using
  genetic programming,'' in \emph{Proc. Int'l Joint Conf. on Computer Vision,
  Imaging and Computer Graphics Theory and Applications}, 2019.

\bibitem[Hayou et~al.(2018)Hayou, Doucet, and Rousseau]{pr:hayou18}
S.~Hayou, A.~Doucet, and J.~Rousseau, ``On the selection of initialization and
  activation function for deep neural networks,'' \emph{arXiv:1805.08266},
  2018.

\bibitem[Li et~al.(2018)Li, Fan, Li, Wu, and Ming]{pr:li18a}
Y.~Li, C.~Fan, Y.~Li, Q.~Wu, and Y.~Ming, ``Improving deep neural network with
  multiple parametric exponential linear units,'' \emph{Neurocomputing}, vol.
  301, pp. 11--24, 2018.

\bibitem[Nair and Hinton(2010)]{pr:nair10}
V.~Nair and G.~E. Hinton, ``Rectified linear units improve restricted boltzmann
  machines,'' in \emph{Proc. Int'l Conf. on Machine Learning}, 2010.

\bibitem[Lin et~al.(2015{\natexlab{a}})Lin, RoyChowdhury, and Maji]{mb:lin15a}
T.-Y. Lin, A.~RoyChowdhury, and S.~Maji, ``Bilinear {CNN} models for
  fine-grained visual recognition,'' in \emph{Proc. Int'l Conf. on Computer
  Vision}, 2015.

\bibitem[Krause et~al.(2014)Krause, Gebru, Deng, Li, and Fei-Fei]{mb:krause14}
J.~Krause, T.~Gebru, J.~Deng, L.-J. Li, and L.~Fei-Fei, ``Learning features and
  parts for fine-grained recognition,'' in \emph{Proc. Int'l Conf. on Pattern
  Recognition}, 2014.

\bibitem[Cohen et~al.(2019)Cohen, Huster, and Cohen]{cohen2019universal}
J.~E. Cohen, T.~Huster, and R.~Cohen, ``Universal lipschitz approximation in
  bounded depth neural networks,'' \emph{arXiv:1904.04861}, 2019.

\bibitem[S{\'a}nchez et~al.(2011)S{\'a}nchez, Perronnin, and
  Akata]{mb:sanchez11}
J.~S{\'a}nchez, F.~Perronnin, and Z.~Akata, ``{Fisher Vectors for Fine-Grained
  Visual Categorization},'' in \emph{Proc. Workshop on Fine-Grained Visual
  Categorization (CVPRW)}, 2011.

\bibitem[Perronnin and Larlus(2015)]{mb:perronnin15}
F.~Perronnin and D.~Larlus, ``Fisher vectors meet neural networks: A hybrid
  classification architecture,'' in \emph{Proc. Conf. on Computer Vision and
  Pattern Recognition}, 2015.

\bibitem[Gao et~al.(2015)Gao, Beijbom, Zhang, and Darrell]{mb:gao15}
Y.~Gao, O.~Beijbom, N.~Zhang, and T.~Darrell, ``Compact bilinear pooling,''
  \emph{CoRR}, vol. abs/1511.06062, 2015.

\bibitem[Kong and Fowlkes(2017)]{mb:kong17}
S.~Kong and C.~Fowlkes, ``Low-rank bilinear pooling for fine-grained
  classification,'' in \emph{Proc. Conf. on Computer Vision and Pattern
  Recognition}, 2017.

\bibitem[Cai et~al.(2017)Cai, Zuo, and Zhang]{mb:cai17}
S.~Cai, W.~Zuo, and L.~Zhang, ``Higher-order integration of hierarchical
  convolutional activations for fine-grained visual categorization,'' in
  \emph{Proc. Int'l Conf. on Computer Vision}, 2017.

\bibitem[Cui et~al.(2017)Cui, Zhou, Wang, Liu, Lin, and Belongie]{mb:cui17}
Y.~Cui, F.~Zhou, J.~Wang, X.~Liu, Y.~Lin, and S.~Belongie, ``Kernel pooling for
  convolutional neural networks,'' in \emph{Proc. Conf. on Computer Vision and
  Pattern Recognition}, 2017.

\bibitem[Zhang et~al.(2014)Zhang, Donahue, Girshick, and Darrell]{mb:zhang14}
N.~Zhang, J.~Donahue, R.~Girshick, and T.~Darrell, ``Part-based {R-CNNs} for
  fine-grained category detection,'' in \emph{Proc. European Conf. on Computer
  Vision}, 2014.

\bibitem[Huang et~al.(2016)Huang, Xu, Tao, and Zhang]{mb:huang16}
S.~Huang, Z.~Xu, D.~Tao, and Y.~Zhang, ``Part-stacked {CNN} for fine-grained
  visual categorization,'' in \emph{Proc. Conf. on Computer Vision and Pattern
  Recognition}, 2016.

\bibitem[Shih et~al.(2015)Shih, Mallya, Singh, and Hoiem]{mb:shi15}
K.~J. Shih, A.~Mallya, S.~Singh, and D.~Hoiem, ``Part localization using
  multi-proposal consensus for fine-grained categorization,'' \emph{CoRR}, vol.
  abs/1507.06332, 2015.

\bibitem[Lin et~al.(2015{\natexlab{b}})Lin, Shen, Lu, and Jia]{mb:lin15b}
D.~Lin, X.~Shen, C.~Lu, and J.~Jia, ``{Deep LAC}: Deep localization, alignment
  and classification for fine-grained recognition,'' in \emph{Proc. Conf. on
  Computer Vision and Pattern Recognition}, 2015.

\bibitem[Simon and Rodner(2015)]{mb:simon15}
M.~Simon and E.~Rodner, ``Neural activation constellations: Unsupervised part
  model discovery with convolutional networks,'' in \emph{Proc. Int'l Conf. on
  Computer Vision}, 2015.

\bibitem[Krause et~al.(2015)Krause, Jin, Yang, and Li]{mb:krause15}
J.~Krause, H.~Jin, J.~Yang, and F.~F. Li, ``Fine-grained recognition without
  part annotations,'' in \emph{Proc. Conf. on Computer Vision and Pattern
  Recognition}, 2015.

\bibitem[Yu et~al.(2018)Yu, Ji, Fu, Guo, Pang, and Zhang]{mb:yu18}
Y.~Yu, Z.~Ji, Y.~Fu, J.~Guo, Y.~Pang, and Z.~Zhang, ``Stacked semantic-guided
  attention model for fine-grained zero-shot learning,'' \emph{CoRR}, vol.
  abs/1805.08113, 2018.

\bibitem[Sun et~al.(2018)Sun, Yuan, Zhou, and Ding]{mb:sun18}
M.~Sun, Y.~Yuan, F.~Zhou, and E.~Ding, ``Multi-attention multi-class constraint
  for fine-grained image recognition,'' in \emph{Proc. European Conf. on
  Computer Vision}, 2018.

\bibitem[Jaderberg et~al.(2015)Jaderberg, Simonyan, Zisserman, and
  Kavukcuoglu]{mb:jaderberg15}
M.~Jaderberg, K.~Simonyan, A.~Zisserman, and K.~Kavukcuoglu, ``Spatial
  transformer networks,'' in \emph{Advances on Neural Information Processing
  Systems}, 2015.

\bibitem[Chen et~al.(2018)Chen, Wu, Gao, Dong, Luo, and Lin]{mb:chen18}
T.~Chen, W.~Wu, Y.~Gao, L.~Dong, X.~Luo, and L.~Lin, ``Fine-grained
  representation learning and recognition by exploiting hierarchical semantic
  embedding,'' in \emph{Proc. ACM Int'l Conf. on Multimedia}, 2018.

\bibitem[Fu et~al.(2017)Fu, Zheng, and Mei]{mb:fu17}
J.~Fu, H.~Zheng, and T.~Mei, ``Look closer to see better: Recurrent attention
  convolutional neural network for fine-grained image recognition,'' in
  \emph{Proc. Conf. on Computer Vision and Pattern Recognition}, 2017.

\bibitem[Wang et~al.(2014)Wang, song, Leung, Rosenberg, Wang, Philbin, Chen,
  and Wu]{mb:wang14}
J.~Wang, Y.~song, T.~Leung, C.~Rosenberg, J.~Wang, J.~Philbin, B.~Chen, and
  Y.~Wu, ``Learning fine-grained image similarity with deep ranking,'' in
  \emph{Proc. Conf. on Computer Vision and Pattern Recognition}, 2014.

\bibitem[Hornik(1991)]{pr:hornik91}
K.~Hornik, ``Approximation capabilities of multilayer feedforward networks,''
  \emph{Neural Networks}, vol.~4, no.~2, pp. 251--257, 1991.

\bibitem[Hochreiter(1998)]{pr:hochreiter98}
S.~Hochreiter, ``The vanishing gradient problem during learning recurrent
  neural nets and problem solutions,'' \emph{Int'l Journal of Uncertainty,
  Fuzziness and Knowledge-Based System}, vol.~6, no.~2, pp. 107--116, 1998.

\bibitem[Maas et~al.(2013)Maas, Hannun, and Ng]{pr:maas13}
A.~L. Maas, A.~Y. Hannun, and A.~Y. Ng, ``Rectifier nonlinearities improve
  neural network acoustic models,'' in \emph{Proc. ICML Workshop on Deep
  Learning for Audio, Speech and Language Processing}, 2013.

\bibitem[Xu et~al.(2015)Xu, Wang, Chen, and Li]{pr:xu15}
B.~Xu, N.~Wang, T.~Chen, and M.~Li, ``Empirical evaluation of rectified
  activations in convolution network,'' \emph{arXiv:1505.00853}, 2015.

\bibitem[He et~al.(2015)He, Zhang, Ren, and Sun]{pr:he15}
K.~He, X.~Zhang, S.~Ren, and J.~Sun, ``Delving deep into rectifiers: Surpassing
  human-level performance on imagenet classification,'' in \emph{Proc. Int'l
  Conf. on Computer Vision}, 2015.

\bibitem[Jin et~al.(2016)Jin, Xu, Feng, Wei, Xiong, and Yan]{Srelu-AF}
X.~Jin, C.~Xu, J.~Feng, Y.~Wei, J.~Xiong, and S.~Yan, ``Deep learning with
  s-shaped rectified linear activation units,'' in \emph{Proceedings {AAAI}
  Conference on Artificial Intelligence}, 2016, pp. 1737--1743.

\bibitem[Trottier et~al.(2017)Trottier, Gigu{\`{e}}re, and
  Chaib{-}draa]{pr:trottier17}
L.~Trottier, P.~Gigu{\`{e}}re, and B.~Chaib{-}draa, ``Parametric exponential
  linear unit for deep convolutional neural networks,'' in \emph{Int'l Conf. on
  Machine Learning and Applications}, 2017.

\bibitem[Szegedy et~al.(2014)Szegedy, Zaremba, Sutskever, Bruna, and
  Dumitru~Erhan]{pr:Szegedy14}
C.~Szegedy, W.~Zaremba, I.~Sutskever, J.~Bruna, and R.~F. Dumitru~Erhan,
  Ian~Goodfellow, ``Intriguing properties of neural networks,'' in \emph{Proc.
  Int'l Conf. on Learning Representations}, 2014.

\bibitem[Weng et~al.(2018)Weng, Zhang, Chen, Song, Hsieh, Daniel, Boning, and
  Dhillon]{TowardsFast}
L.~Weng, H.~Zhang, H.~Chen, Z.~Song, C.-J. Hsieh, L.~Daniel, D.~Boning, and
  I.~Dhillon, ``Towards fast computation of certified robustness for {R}e{LU}
  networks,'' in \emph{Proc. Int'l Conf. on Machine Learning}, 2018.

\bibitem[Scaman and Virmaux(2018)]{lipchitzregualrity}
K.~Scaman and A.~Virmaux, ``Lipschitz regularity of deep neural networks:
  Analysis and efficient estimation,'' in \emph{Advances on Neural Information
  Processing Systems}, 2018.

\bibitem[Fazlyab et~al.(2019)Fazlyab, Robey, Hassani, Morari, and
  Pappas]{fazlyab}
M.~Fazlyab, A.~Robey, H.~Hassani, M.~Morari, and G.~J. Pappas, ``Efficient and
  accurate estimation of lipschitz constants for deep neural networks,''
  \emph{arXiv:1906.04893}, 2019.

\bibitem[Sokoli\'{c} et~al.(2017)Sokoli\'{c}, Giryes, Sapiro, and
  Rodrigues]{largemargin}
J.~Sokoli\'{c}, R.~Giryes, G.~Sapiro, and M.~Rodrigues, ``Robust large margin
  deep neural networks,'' \emph{IEEE Transactions on Signal Processing},
  vol.~65, no.~16, pp. 4265--4280, 2017.

\bibitem[Anil et~al.(2019)Anil, Lucas, and Grosse]{sorting}
C.~Anil, J.~Lucas, and R.~B. Grosse, ``Sorting out lipschitz function
  approximation,'' in \emph{Proc. Int'l Conf. on Machine Learning}, 2019.

\bibitem[Qian and Wegman(2019)]{l2nonexpansive}
H.~Qian and M.~N. Wegman, ``L2-nonexpansive neural networks,'' in \emph{Proc.
  Int'l Conf. on Learning Representations}, 2019.

\bibitem[Tsuzuku et~al.(2018)Tsuzuku, Sato, and Sugiyama]{lipschitzmargin}
Y.~Tsuzuku, I.~Sato, and M.~Sugiyama, ``Lipschitz-margin training: Scalable
  certification of perturbation invariance for deep neural networks,'' in
  \emph{Advances on Neural Information Processing Systems}, 2018.

\bibitem[Gordon(1995)]{pr:gordon95}
G.~J. Gordon, ``Stable function approximation in dynamic programming,'' in
  \emph{Proc. Int'l Conf. on Machine Learning}, 1995.

\bibitem[Sagan(1974)]{pr:sagan74}
H.~Sagan, \emph{Advanced Calculus}.\hskip 1em plus 0.5em minus 0.4em\relax
  Houghton Mifflin, 1974.

\bibitem[Krizhevsky(2009)]{pr:krishevski09}
A.~Krizhevsky, ``Learning multiple layers of features from tiny images,''
  University of Toronto, Tech. Rep. 1 (4), 7, 2009.

\bibitem[Wah et~al.(2011)Wah, Branson, Welinder, Perona, and
  Belongie]{mb:wah11}
C.~Wah, S.~Branson, P.~Welinder, P.~Perona, and Belongie, ``{The Caltech-UCSD
  Birds-200-2011 Dataset},'' California Institute of Technology, Tech. Rep.
  CNS-TR-2011-001, 2011.

\bibitem[Krause et~al.(2013)Krause, Stark, Deng, and Fei-Fei]{mb:krause13}
J.~Krause, M.~Stark, J.~Deng, and L.~Fei-Fei, ``{3D} object representations for
  fine-grained categorization,'' in \emph{Proc. Int'l Workshop on 3D
  Representation and Recognition}, 2013.

\bibitem[Khosla et~al.(2011)Khosla, Jayadevaprakash, Yao, and
  Fei-Fei]{mb:khosla11}
A.~Khosla, N.~Jayadevaprakash, B.~Yao, and L.~Fei-Fei, ``Novel dataset for
  fine-grained image categorization,'' in \emph{Proc. Workshop on Fine-Grained
  Visual Categorization (CVPRW)}, 2011.

\bibitem[Maji et~al.(2013)Maji, Rahtu, Kannala, Blaschko, and
  Vedaldi]{mb:maji13}
S.~Maji, E.~Rahtu, J.~Kannala, M.~Blaschko, and A.~Vedaldi, ``Fine-grained
  visual classification of aircraft,'' \emph{arXiv:1306.5151}, 2013.

\bibitem[Kaggle()]{mb:iFood}
Kaggle, ``\url{https://sites.google.com/view/fgvc5/competitions/fgvcx/ifood},''
  (last accessed: July 25, 2019).

\end{thebibliography}
